# Interpretable Predictive Models for Healthcare via Rational Logistic Regression

*Completed Research Paper*

**Thiti Suttaket**
Department of Information Systems
and Analytics
National University of Singapore
e0384140@u.nus.edu

**L Vivek Harsha Vardhan**
Department of Information Systems
and Analytics
National University of Singapore
harsha@comp.nus.edu.sg

**Stanley Kok**
Department of Information Systems and Analytics
National University of Singapore
skok@comp.nus.edu.sg

## Abstract

*The healthcare sector has experienced a rapid accumulation of digital data recently, especially in the form of electronic health records (EHRs). EHRs constitute a precious resource that IS researchers could utilize for clinical applications (e.g., morbidity prediction). Deep learning seems like the obvious choice to exploit this surfeit of data. However, numerous studies have shown that deep learning does not enjoy the same kind of success on EHR data as it has in other domains; simple models like logistic regression are frequently as good as sophisticated deep learning ones. Inspired by this observation, we develop a novel model called rational logistic regression (RLR) that has standard logistic regression (LR) as its special case (and thus inherits LR's inductive bias that aligns with EHR data). RLR has rational series as its theoretical underpinnings, works on longitudinal time-series data, and learns interpretable patterns. Empirical comparisons on real-world clinical tasks demonstrate RLR's efficacy.*

## Introduction

The widespread adoption of information systems in past decades, coupled with its associated digitalization of all kinds of media, has led to the accumulation of a large amount of data, a precious resource that can potentially serve as fodder for machine learning systems. Using this preponderance of data as training inputs, deep learning (Goodfellow et al. 2016) has recently demonstrated groundbreaking empirical results in a wide variety of domains, ranging from machine translation (Luong et al. 2015) and speech recognition (Synnaeve et al. 2020) to advertising (Badhe 2015) and e-commerce (Tan et al. 2020).

The healthcare domain has similarly experienced a rapid growth in the amount of digital data collected, especially in the form of patient electronic health records (EHRs). EHR systems in hospitals now routinely record all patient-related information (e.g., demographic details, physiological measurements, diagnoses, medications, laboratory tests, surgical procedures, and visit dates) (Birkhead et al. 2015). While the main purpose of EHR systems continues to be the facilitation of day-to-day healthcare operations, they present an opportunity for IS researchers to utilize their patient-specific data for various secondary clinical applications (e.g., disease prediction (Austin et al. 2013) and patient stratification (Doshi-Velez et al. 2014)). However, unlike in other domains, deep learning has largely *not* enjoyed the same kinds of empirical breakthroughs with EHR data. The simple machine learning workhorse, (regularized) logistic regression frequently performs on par with sophisticated deep learning models on a variety of clinical prediction problems (Rajkomar et al. 2018)[1]. Recent research (Christodoulou et al. 2019) also finds no evidence suggesting that deep learning is superior over logistic regression for clinical prediction modeling.

The reasons for deep learning's unexpectedly lackluster performance (and logistic regression's surprisingly good results) may lie in three characteristics of EHR data.

First, many components of EHR data are individually information-rich, and represent decades of clinical research into their usefulness. For example, the laboratory test for the level of HbA1c protein in a patient's blood is frequently used as an indicator of blood glucose level, and its efficacy as a biomarker for diabetes is backed by years of medical research. Thus, deep learning's ability to derive good features from raw input data may be less important for EHR data that already contains informative features to begin with[2].

Second, another characteristic of EHR data that stymies deep learning is its relatively small data size. Electronic databases of hospitals typically contain records of only a few hundred thousand patients. This number, though seemingly large, is minuscule compared to the data present in domains where deep learning excels (e.g., neural machine translation typically has tens of millions of source-target sentence pairs as training examples). This problem is compounded by the even smaller number of patients with a particular morbidity. For instance, while there may be hundreds of thousands of patients in an EHR database, only a small percentage of them are afflicted by a disease of interest (see the *Experiment* section for concrete examples).

Third, EHR data has widely varying time-scales (e.g., the blood pressure of an intensive-care patient could be taken hourly, whereas medications of chronic-diseased patients could be prescribed monthly). This wide disparity of time scales differs significantly from domains where deep learning is highly successful (e.g., machine translation proceeds at one-word-per-time-step intervals, and speech recognition samples data at regular intervals of several milliseconds).

Note that even if deep learning is successful on EHR data, its impenetrable "black-box" nature prevents medical practitioners from elucidating how its predictions are made, and hence hinders its adoption in clinical settings where decisions have to be justified by explanations. Some efforts have been made to ameliorate this issue via attention mechanisms (Choi et al. 2016a; Lee et al. 2018; Ma et al. 2017; Ma et al. 2020), which weight inputs at each time point in order to highlight those that are most relevant to a prediction task. However, only highlighting which inputs are important without sufficient clarity on how they correlate with each other and with the prediction task causes end-users to be confused (Payrovnaziri et al. 2020), and demotes trust among them towards machine learning models. To compound the problem, the attention weights are frequently so diffused that a multitude of inputs are simultaneously highlighted across time points. Not only would this obfuscate the identities of the truly important inputs, but would also result in information overload for the end-users. An alternative approach for providing some interpretability to deep learning models uses post-hoc explanations. Such an approach (e.g., LIME (Ribeiro et al. 2016)) is model-agnostic, i.e., it can be applied to models other than deep learning ones. It involves learning another interpretable model (usually a simple one like a decision tree) from the predictions of the blackbox model of interest so as to mimic the blackbox model. The ability of the interpretable model to represent the blackbox model is called *fidelity* or *faithfulness*. However, if the decision function of the blackbox model is overly complex, the simple model may be ill-suited to imitate the blackbox model well enough in terms of the predictions made (i.e., the simple model has low fidelity to the blackbox model).

On the other hand, the good performance of logistic regression suggests that it has a good *inductive bias*

---

[1]See Table 1 sequestered on the fourth last page of the paper's supplementary material.
[2]One may regard these informative clinical features to have been condensed from years of learning by deep learning's biological counterpart – the (collective) human brain.

("fit") for EHR data. This apposite inductive bias, in tandem with its simplicity, allows logistic regression to achieve a good balance between accuracy and interpretability. This is in contrast to the interpretability-accuracy trade-off commonly faced by machine learning models (Bohanec and Bratko 1994), where a simple, interpretable model lacks the necessary accuracy, while a more accurate model is too complicated and inscrutable.

However, logistic regression has a drawback of not being able to work directly with longitudinal EHR data; it has to first summarize sequential EHR data into aggregate features before using them as inputs, thereby discarding potentially useful temporal information in the data.

In this paper, we propose a model called *rational* logistic regression (RLR) that has vanilla logistic regression as its special case (and hence inherits its useful inductive bias), and generalizes it so as to model the temporal and sequential information present in longitudinal data too. RLR is built upon *weighted finite state automata* (WFSAs) and their associated concept of *rational* languages (see Background section). By inheriting the strong inductive bias of logistic regression, RLR continues to be amenable to the features in EHR data. By incorporating the constraints implicit in the inductive bias, RLR requires less data to train than deep learning systems (that do not have such constraints to guide their learning). Further, RLR learns human-intepretable *sequential* patterns, each representing the progression of (a combination of) biomarkers of a disease.

Aside from capturing patterns at the population level, RLR can provide *personalized* explanations for an individual patient by highlighting the patterns that contributed the most to a prediction for that patient. Because each pattern can learn to ignore unimportant information at varying time intervals, it can flexibly focus on salient features at multiple time scales. These characteristics of RLR are direct counterpoints to the aforementioned problems associated with deep learning for EHR data.

In sum, our contributions are as follows.

- We propose a novel, data-efficient, interpretable model called RLR for predictive healthcare tasks.
- We rigorously show how RLR is a generalization of logistic regression to sequential data.
- We empirically compare our RLR model to state-of-the-art baselines, and show that RLR outperforms them in terms of predictive accuracy on 4 real-world medical tasks.
- We qualitatively demonstrate the interpretability of the patterns that RLR learns, and describe the "reasoning" that is illustrated by each pattern.

## Related Work

Being able to make accurate predictions for clinical events (such as mortality or disease onset) is an important problem because of its potential to improve healthcare outcomes and reduce operational cost. The traditional approaches to tackle this problem are largely based on disease severity classification systems. In such systems, a list of criteria indicative of a disease (or clinical event such as death) is identified, and a score is assigned based on the number of criteria that a patient satisfies. SAPS (Bisbal et al. 2014) is a classification system used to evaluate the mortality risk for post-cardiac arrest patients. APACHE (Zimmerman et al. 2006) has been developed to quantify mortality risk in intensive-care units using clinical factors such as care procedures administered and medications prescribed. LACE (Van Walraven et al. 2010) evaluates the risk of readmission, and death after discharge from a hospital. HOSPITAL Score (Donzé et al. 2013) is used to evaluate the risk of 30-day readmission.

More recently, sophisticated deep learning approaches have been applied to clinical prediction. Choi et al. 2016b use recurrent neural networks (RNNs) to predict in-patient mortality upon hospital admission using historical medical records. Che et al. 2018 improve upon this by imputing missing data. Ma et al. 2017 use a more sophisticated deep learning model based on bidirectional RNNs to make predictions. However, it is not necessarily the case that deep learning approaches are superior to simple models such as logistic regression (Rajkomar et al. 2018, Supplemental Table 1). Christodoulou et al. 2019 investigate the performance of deep learning approaches vis-a-vis logistic regression in clinical prediction modeling using 282 previous studies, and found no evidence suggesting that deep learning approaches are superior to logistic regression.

Inspired by this observation, Kodialam et al. 2021 use logistic regression to guide the training of a deep learning model (in a process called reverse distillation).

Attention mechanisms in deep learning (Lee et al. 2018; Ma et al. 2017) have been used to measure the relationships among different visits. Sun et al. 2018 highlight the relative importance of clinical features via adversarial attack in a deep learning framework. RETAIN (Choi et al. 2016a) is a state-of-the-art deep learning system that uses a two-level neural attention mechanism upon an RNN to model longitudinal EHR data. To generate attention weights for interpretability, RETAIN uses two RNNs. The first RNN (RNN $\alpha$) uses the softmax function to get visit-level attention weights, and the second RNN (RNN $\beta$) uses the tanh function to calculate clinical-variable-level attention weights. Adacare (Ma et al. 2020) is another state-of-the-art deep learning approach that also uses an attention mechanism for interpretability. In addition, it uses dilated convolutions with multi-scale receptive fields to capture the long and short-term historical variations in longitudinal EHR data.

Both RETAIN and Adacare take a step towards interpretability through their use of neural attention mechanisms. They are able to identify important features at each time step across a time span. However, they do not elucidate the connections among the features at different time steps. In contrast, our RLR model learns interpretable patterns that thread together features at different time steps into a coherent whole. For example, RETAIN's and Adacare's attention mechanism can identify important features $A_1$ and $B_1$ at time-step 1 and important features $A'_2$ and $B'_2$ at time step 2. However, they do not know how a feature at time-step 1 cohere with a feature at time-step 2. Does $A_1 \rightarrow A'_2$ ($A'_2$ follow $A_1$)? Or $A_1 \rightarrow B'_2$? Or do combination of features interact, e.g., $(A_1, B_1) \rightarrow B'_2$ and $B_1 \rightarrow (A'_2, B'_2)$? Or perhaps all of the above? Our RLR model can learn patterns to coherently capture all such regularities.

To handle the multi-time-scale nature of longitudinal EHR data, Adacare takes a brute force approach. It creates convolutional filters that skip inputs at regular intervals of $k$ time steps, i.e., each filter considers inputs separated by exactly $k$ time steps ($k = 1, 2, \ldots, 5$ in its experiments). In contrast, our RLR model finesses the problem by allowing patterns to flexibly learn time spans over which to ignore inputs, and different such time spans can appear in a single pattern.

## Background

Our model RLR is built upon semirings and weighted finite state automata. We describe each of them in turn.

### Semirings

A *semiring* (Kuich and Salomaa 1986) is defined as a set $\mathbb{K}$ that is associated with two binary operations, $\oplus$ (addition) and $\otimes$ (multiplication), and two identity elements, $\bar{0}$ (for addition) and $\bar{1}$ (for multiplication). A semiring requires both $\oplus$ and $\otimes$ to be associative, $\oplus$ to be commutative, $\otimes$ to distribute over addition, and $\bar{0}$ to annihilate under $\otimes$ (i.e., $\bar{0} \otimes k = k \otimes \bar{0} = \bar{0}, \forall k \in \mathbb{K}$). A semiring can be specified in terms of a 5-tuple corresponding to $(\mathbb{K}, \oplus, \bar{0}, \otimes, \bar{1})$. The most common semiring is plus-times $((\mathbb{R}, +, 0, \times, 1))$; other common examples of semirings are:

- min-plus $((\mathbb{R} \cup \{\infty\}, \min, \infty, +, 0))$,
- max-plus $((\mathbb{R} \cup \{-\infty\}, \max, -\infty, +, 0))$,
- min-product $((\mathbb{R}_{>0} \cup \{\infty\}, \min, \infty, \times, 1))$, and
- max-product $((\mathbb{R}_{\geq 0}, \max, 0, \times, 1))$.

As shown later, our RLR model utilizes the max-product semiring.

### Weighted Finite State Automaton (WFSA)

WFSAs (Sipser 1986) are commonly used in natural language processing to model text strings, and we use a simple example in that domain to provide an intuitive understanding. In Figure 1, an initial state is represented by a bold circle, and a final state is represented by a double circle. A transition from one state to another is represented by an arc from the former to the latter. Each arc is labeled with the input symbol (e.g., 'good') and weight (e.g., 0.7) associated with the transition. The transition weight is typically interpreted as a (unnormalized) probability. Beginning from the initial state, we traverse consecutive arcs, each time *consuming* the symbol associated with the arc, and accumulating its weight. In this manner, a WFSA provides a mapping from a sequence of symbols (e.g., words) to a sequence of (scalar) weights. A transition labeled with the *empty symbol* $\epsilon$ consumes no input. A path is a sequence of consecutive transitions, and a *successful* path is one that starts in the initial state and ends in the final state. The label of a path is the concatenation of the labels of its constituent transitions. The weight of the path is the product of an initial weight, the weights of its constituent transitions, and the weight of the final state reached by the path.

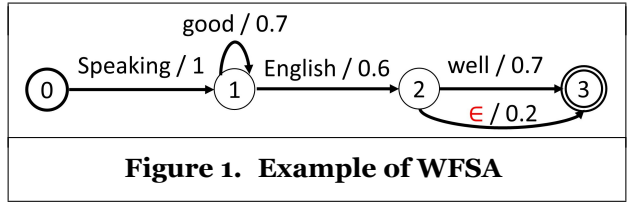

**Figure 1. Example of WFSA**

An input sequence $x_1, \ldots, x_n$ is termed to be *accepted* by a WFSA if there exists a successful path in the WFSA labeled with $x_1, \ldots, x_n$ (modulo the $\epsilon$ symbol). For example, the WFSA in Figure 1 accepts "Speaking English well" and "Speaking good English", but not "Speaking good" and "Speaking Spanish and English". The weight associated with an input sequence $x_1, \ldots, x_n$, also known as its *score*, is the sum of the weights of all possible successful paths labeled $x_1, \ldots, x_n$ (modulo the $\epsilon$ symbol). Intuitively, the WFSA is matching input sequences to a pattern, only accepting those that conform to the pattern.

Formally, a WFSA is associated with a semiring $(\mathbb{K}, \oplus, \bar{0}, \otimes, \bar{1})$ that defines its addition and multiplication operators. A WFSA is defined as a 5-tuple $(V, Q, \mathbf{T}, \pi, \eta)$, where $V$ is the input set, $Q$ is a set of states with size $|Q| = d$, $\pi \in \mathbb{K}^{1 \times d}$ is an initial weight row vector, $\eta \in \mathbb{K}^{d \times 1}$ is a final weight column vector, and $\mathbf{T} : (V \cup \{\epsilon\})$ is a transition weight function. $\pi$ and $\eta$ give the probabilities of starting and ending in each state respectively. $\mathbf{T}(\cdot)$ can be interpreted as a $d \times d$ matrix. ($\mathbf{T}$ need not be constrained to take a discrete symbol as input; as will be shown later, it can also accept a real-valued vector as input.)

Given a sequence of inputs $\mathbf{x} = (x_1, \ldots, x_n) \in V^n$, the Forward algorithm (Baum and Petrie 1966) scores $\mathbf{x}$ with respect to a WFSA. In the absence of $\epsilon$-transitions, Forward expresses the score $p'_{span}(\mathbf{x})$ as a series of matrix multiplications:

$$p'_{span}(\mathbf{x}) = \pi \left( \Pi_{i=1}^n \mathbf{T}(x_i) \right) \eta. \tag{1}$$

Since an $\epsilon$-transition occurs without consuming an input symbol $x_i \in V$, to incorporate $\epsilon$-transitions, we can rewrite Equation 1 as

$$
\begin{aligned}
p_{span}(\mathbf{x}) &= \pi \mathbf{T}(\epsilon)^* \left( \Pi_{i=1}^n \mathbf{T}(x_i) \mathbf{T}(\epsilon)^* \right) \eta \\
&= \left( \left( \left( \ldots \left( \left( (\pi \ \mathbf{T}(\epsilon)^*) \ \mathbf{T}(x_1) \right) \mathbf{T}(\epsilon)^* \ldots \ \mathbf{T}(x_n) \right) \mathbf{T}(\epsilon)^* \right) \eta \right)
\end{aligned} \tag{2}
$$

where $*$ refers to matrix asteration ($\mathbf{A}^* := \sum_{j=0}^\infty \mathbf{A}^j$). Equation 2 can be rewritten recursively as

$$\mathbf{h}_0 = \pi \ \mathbf{T}(\epsilon)^* \tag{3}$$

$$\mathbf{h}_{t+1} = \left( \mathbf{h}_t \ \mathbf{T}(x_{t+1}) \right) \mathbf{T}(\epsilon)^* \tag{4}$$

$$p_{span}(\mathbf{x}) = \mathbf{h}_n \ \eta \tag{5}$$

### Neural WFSA

Schwartz et al. 2018 created a neural version of WFSA by using a neural network to learn the transition weight function $\mathbf{T}(\cdot)$. For tractability, they also restricted transitions from each state $i$ to the following 3 kinds.

1. *Self-loop*: The transition consumes an input symbol and stays at the same state $i$.

2. *Main path*: The transition consumes an input symbol and moves to state $i+1$. (NB: Moving to state $i + k$ where $k \neq 1$ is not allowed.)

3. *$\epsilon$-transition*: The transition moves to state $i + 1$ without consuming an input symbol.

(Note that the self-loop transition and $\epsilon$-transition allow a WFSA to accept an input sequence that is respectively longer and shorter than its number of states.)

The transition weight function $\mathbf{T}(\cdot)$ is defined as

$$[\mathbf{T}(x)]_{i,j} = \begin{cases} E(\mathbf{u}_i \cdot x + a_i) & \text{if } j = i \text{ (self-loop)} \\ E(\mathbf{w}_i \cdot x + b_i) & \text{if } j = i + 1 \\ 0 & \text{otherwise} \end{cases} \tag{6}$$

where $x$ is now an input real-valued *vector*, $[\mathbf{T}(x)]_{i,j}$ is the element at $(i, j)$ position of matrix $\mathbf{T}(x)$, $\mathbf{u}_i$ and $\mathbf{w}_i$ are vectors of parameters (*weights*), $a_i$ and $b_i$ are scalar parameters (*biases*), and $E$ is an encoding function. Because $x$ is a vector (rather than a discrete symbol), a neural WFSA can flexibly accept a vector of real-valued features at each time step. For $\epsilon$-transitions, $\mathbf{T}(\epsilon)$ is defined as

$$[\mathbf{T}(\epsilon)]_{i,j} = \begin{cases} E(c_i) & \text{if } j = i + 1 \\ 0 & \text{otherwise} \end{cases} \tag{7}$$

where $c_i$ is a scalar parameter.

The score of an input sequence $p_{span}(\mathbf{x})$ is computed recursively in the same way as for a vanilla WFSA using Equations 3-5. However, for tractability, $\mathbf{T}(\epsilon)^*$ in those equations is approximated as $\mathbf{I} + \mathbf{T}(\epsilon)$. In a neural WFSA, the addition and multiplication operators are also defined by a semiring.

Peng et al. 2018 defines the sets of strings accepted by neural WFSAs as *rational* languages[3]. Since our model RLR is built upon neural WFSAs, we mnemonically name it *rational* logistic regression to reflect its ability to learn patterns that are rational languages.

## Rational Logistic Regression (RLR)

To create our RLR model, we adopt the neural WFSA framework defined by Schwartz et al. 2018, and instantiate WFSAs by setting the encoding function $E(\cdot)$ in Equations 6 and 7 to the sigmoid function (i.e., $E(\cdot) = \sigma(\cdot)$), and use the max-product semiring ($(\mathbb{R}_{\geq 0}, \max, 0, \times, 1)$). Our RLR model thus created is a WFSA.

First we show that simple logistic regression is equivalent to an RLR with 2 states (0 and 1) and no $\epsilon$-transition (achieved by setting $T(\epsilon)^*$ to the identity matrix). In such an RLR, the transition weight functions (Equations 6 and 7), and initial and final weight vectors are

$$\mathbf{T}(x) = \begin{bmatrix} \sigma(\mathbf{u}_0 \cdot x + a_0) & \sigma(\mathbf{w}_0 \cdot x + b_0) \\ 0 & \sigma(\mathbf{u}_1 \cdot x + a_1) \end{bmatrix}, \quad \mathbf{T}(\epsilon)^* = \begin{bmatrix} 1 & 0 \\ 0 & 1 \end{bmatrix}, \quad \pi = \begin{bmatrix} 1 & 0 \end{bmatrix}, \quad \eta = \begin{bmatrix} 0 \\ 1 \end{bmatrix}.$$

Using Equation 2, we get the score for input $x_1$ (a real-valued vector) as

$$p_{span}(x_1) = \begin{bmatrix} 1 & 0 \end{bmatrix} \begin{bmatrix} 1 & 0 \\ 0 & 1 \end{bmatrix} \begin{bmatrix} \sigma(\mathbf{u}_0 \cdot x_1 + a_0) & \sigma(\mathbf{w}_0 \cdot x_1 + b_0) \\ 0 & \sigma(\mathbf{u}_1 \cdot x_1 + a_1) \end{bmatrix} \begin{bmatrix} 1 & 0 \\ 0 & 1 \end{bmatrix} \begin{bmatrix} 0 \\ 1 \end{bmatrix} = \sigma(\mathbf{w}_0 \cdot x_1 + b_0),$$

which is *exactly* the equation of a logistic regression model. (NB: Because we are using the max-product semiring, all sum operations resulting from the matrix multiplications in the above expression must be replaced with the max operation. This also applies to the matrix multiplications in subsequent equations.)

Next we illustrate with concrete examples that RLR with more than 2 states is equivalent to accumulating the outputs of logistic regression functions over individual inputs in a sequence. For simplicity, we focus on an RLR with 3 states (depicted in Figure 2(a)) with no $\epsilon$-transitions (we omit the identity $\mathbf{T}(\epsilon)^*$ matrix in the multiplications below to save space). For such an RLR, we have

$$\mathbf{T}(x) = \begin{bmatrix} \sigma(\mathbf{u}_0 \cdot x + a_0) & \sigma(\mathbf{w}_0 \cdot x + b_0) & 0 \\ 0 & \sigma(\mathbf{u}_1 \cdot x + a_1) & \sigma(\mathbf{w}_1 \cdot x + b_1) \\ 0 & 0 & \sigma(\mathbf{u}_2 \cdot x + b_2) \end{bmatrix}, \mathbf{T}(\epsilon)^* = \begin{bmatrix} 1 & 0 & 0 \\ 0 & 1 & 0 \\ 0 & 0 & 1 \end{bmatrix}, \pi = \begin{bmatrix} 1 & 0 & 0 \end{bmatrix}, \eta = \begin{bmatrix} 0 \\ 0 \\ 1 \end{bmatrix}. \quad (8)$$

---

[3]Rational languages are analogous to *regular* languages that are accepted by *unweighted* finite state automata.

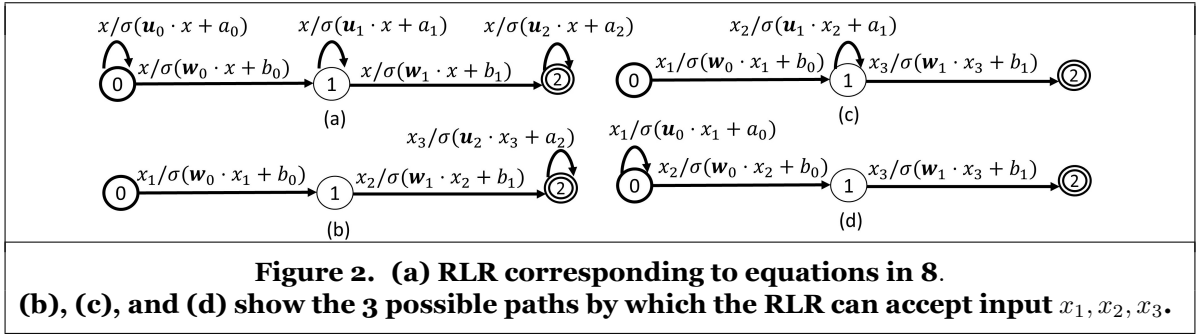

**Figure 2.  (a) RLR corresponding to equations in 8.**
**(b), (c), and (d) show the 3 possible paths by which the RLR can accept input** $x_1, x_2, x_3$.

As before, using Equation 2, we get the score for input sequence $x_1, x_2$ as

$$p_{span}(x_1, x_2) = \begin{bmatrix} 1 & 0 & 0 \end{bmatrix} \begin{bmatrix} \sigma(\mathbf{u}_0 \cdot x_1 + a_0) & \sigma(\mathbf{w}_0 \cdot x_1 + b_0) & 0 \\ 0 & \sigma(\mathbf{u}_1 \cdot x_1 + a_1) & \sigma(\mathbf{w}_1 \cdot x_1 + b_1) \\ 0 & 0 & \sigma(\mathbf{u}_2 \cdot x_1 + b_2) \end{bmatrix}$$

$$\begin{bmatrix} \sigma(\mathbf{u}_0 \cdot x_2 + a_0) & \sigma(\mathbf{w}_0 \cdot x_2 + b_0) & 0 \\ 0 & \sigma(\mathbf{u}_1 \cdot x_2 + a_1) & \sigma(\mathbf{w}_1 \cdot x_2 + b_1) \\ 0 & 0 & \sigma(\mathbf{u}_2 \cdot x_2 + b_2) \end{bmatrix} \begin{bmatrix} 0 \\ 0 \\ 1 \end{bmatrix}$$

$$= \sigma(\mathbf{w}_0 \cdot x_1 + b_0) \, \sigma(\mathbf{w}_1 \cdot x_2 + b_1). \quad (9)$$

This is equivalent to accumulating logistic regression at each transition for an input $x_i$. Longer sequences such as $x_1, x_2, x_3$ can also be matched with this RLR by using self-loop transitions.

$$p_{span}(x_1, x_2, x_3) = \begin{bmatrix} 1 & 0 & 0 \end{bmatrix} \begin{bmatrix} \sigma(\mathbf{u}_0 \cdot x_1 + a_0) & \sigma(\mathbf{w}_0 \cdot x_1 + b_0) & 0 \\ 0 & \sigma(\mathbf{u}_1 \cdot x_1 + a_1) & \sigma(\mathbf{w}_1 \cdot x_1 + b_1) \\ 0 & 0 & \sigma(\mathbf{u}_2 \cdot x_1 + b_2) \end{bmatrix}$$

$$\begin{bmatrix} \sigma(\mathbf{u}_0 \cdot x_2 + a_0) & \sigma(\mathbf{w}_0 \cdot x_2 + b_0) & 0 \\ 0 & \sigma(\mathbf{u}_1 \cdot x_2 + a_1) & \sigma(\mathbf{w}_1 \cdot x_2 + b_1) \\ 0 & 0 & \sigma(\mathbf{u}_2 \cdot x_2 + b_2) \end{bmatrix}$$

$$\begin{bmatrix} \sigma(\mathbf{u}_0 \cdot x_3 + a_0) & \sigma(\mathbf{w}_0 \cdot x_3 + b_0) & 0 \\ 0 & \sigma(\mathbf{u}_1 \cdot x_3 + a_1) & \sigma(\mathbf{w}_1 \cdot x_3 + b_1) \\ 0 & 0 & \sigma(\mathbf{u}_2 \cdot x_3 + b_2) \end{bmatrix} \begin{bmatrix} 0 \\ 0 \\ 1 \end{bmatrix}$$

$$= \max(\sigma(\mathbf{w}_0 \cdot x_1 + b_0)\sigma(\mathbf{w}_1 \cdot x_2 + b_1)\sigma(\mathbf{u}_2 \cdot x_3 + a_2),$$
$$\sigma(\mathbf{w}_0 \cdot x_1 + b_0)\sigma(\mathbf{u}_1 \cdot x_2 + a_1)\sigma(\mathbf{w}_1 \cdot x_3 + b_1),$$
$$\sigma(\mathbf{u}_0 \cdot x_1 + a_0)\sigma(\mathbf{w}_0 \cdot x_2 + b_0)\sigma(\mathbf{w}_1 \cdot x_3 + b_1)), \quad (10)$$

The score of sequence $x_1, x_2, x_3$ is given by the maximum among:

1. $\sigma(\mathbf{w}_0 \cdot x_1 + b_0)\sigma(\mathbf{w}_1 \cdot x_2 + b_1)\sigma(\mathbf{u}_2 \cdot x_3 + a_2)$, obtained by transitioning from state 0 to 1 by input $x_1$, from state 1 to 2 by $x_2$, and self-looping at state 2 by $x_3$ (see Figure 2(b));

2. $\sigma(\mathbf{w}_0 \cdot x_1 + b_0)\sigma(\mathbf{u}_1 \cdot x_2 + a_1)\sigma(\mathbf{w}_1 \cdot x_3 + b_1)$, obtained by transitioning from state 0 to 1 by input $x_1$, self-looping at state 1 by input $x_2$, and transitioning from state 1 to state 2 by $x_3$ (Figure 2(c));

3. $\sigma(\mathbf{u}_0 \cdot x_1 + a_0)\sigma(\mathbf{w}_0 \cdot x_2 + b_0)\sigma(\mathbf{w}_1 \cdot x_3 + b_1)$, obtained by self-looping at state 0 by input $x_1$, transitioning from state 0 to 1 by $x_2$, and transitioning from state 1 to 2 by $x_3$ (Figure 2(d)).

Each of these three cases involves the multiplication of logistic regression functions, with each function applied over individual inputs in a sequence and for different transitions.

Till now we have disallowed $\epsilon$-transitions. Recall from the *Neural WFSA* section that $\mathbf{T}(\epsilon)^* \approx \mathbf{I} + \mathbf{T}(\epsilon) = \max(\mathbf{I}, \mathbf{T}(\epsilon))$ (RLR uses the max-product semiring that uses max as its addition operator). Thus, $\mathbf{T}(\epsilon)^*$ is as follows (the resulting RLR is shown in Figure 3(a)).

$$\mathbf{T}(\epsilon)^* = \begin{bmatrix} 1 & \sigma(c_0) & 0 \\ 0 & 1 & \sigma(c_1) \\ 0 & 0 & 1 \end{bmatrix}.$$

We shall see below that $\epsilon$-transitions allow RLR to accept a sequence $(x_1)$ whose length is shorter than the number of main path transitions needed to reach the end state (2 in this case). The score of $x_1$ is

$$p_{span}(x_1) = \begin{bmatrix} 1 & 0 & 0 \end{bmatrix} \begin{bmatrix} 1 & \sigma(c_0) & 0 \\ 0 & 1 & \sigma(c_1) \\ 0 & 0 & 1 \end{bmatrix} \begin{bmatrix} \sigma(\mathbf{u}_0 \cdot x_1 + a_0) & \sigma(\mathbf{w}_0 \cdot x_1 + b_0) & 0 \\ 0 & \sigma(\mathbf{u}_1 \cdot x_1 + a_1) & \sigma(\mathbf{w}_1 \cdot x_1 + b_1) \\ 0 & 0 & \sigma(\mathbf{u}_2 \cdot x_1 + a_2) \end{bmatrix}$$
$$\begin{bmatrix} 1 & \sigma(c_0) & 0 \\ 0 & 1 & \sigma(c_1) \\ 0 & 0 & 1 \end{bmatrix} \begin{bmatrix} 0 \\ 0 \\ 1 \end{bmatrix} = \begin{matrix} \max(\sigma(\mathbf{w}_0 \cdot x_1 + b_0)\sigma(c_1), \\ \sigma(c_0)\sigma(\mathbf{u}_1 \cdot x_1 + a_1)\sigma(c_1), \\ \sigma(c_0)\sigma(\mathbf{w}_1 \cdot x_1 + b_1)). \end{matrix}$$

This shows that the sequence with length 1 can be accepted by the RLR with 3 states and its score is equal to the maximum of

1. $\sigma(\mathbf{w}_0 \cdot x_1 + b_0)\sigma(c_1)$, obtained by transitioning from state 0 to 1 by input $x_1$, and transitioning from state 1 to 2 by an $\epsilon$-transition (Figure 3(b)).

2. $\sigma(c_0)\sigma(\mathbf{u}_1 \cdot x_1 + a_1)\sigma(c_1)$, obtained by transitioning from state 0 to 1 by an $\epsilon$-transition, self-looping at state 1 by $x_1$, and transitioning from state 1 to 2 by an $\epsilon$-transition (Figure 3(c)).

3. $\sigma(c_0)\sigma(\mathbf{w}_1 \cdot x_1 + b_1)$, obtained by transitioning from state 0 to 1 by an $\epsilon$-transition, and transitioning from state 1 to 2 by input $x_1$ (Figure 3(d)).

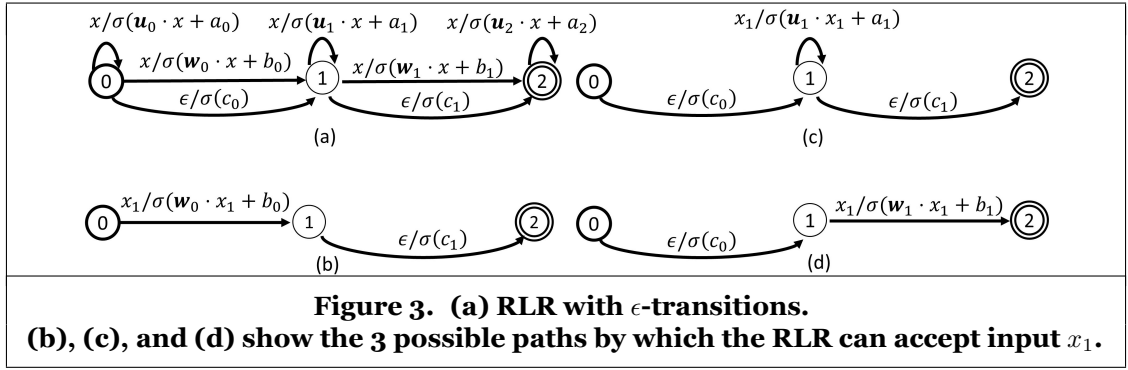

**Figure 3. (a) RLR with $\epsilon$-transitions. (b), (c), and (d) show the 3 possible paths by which the RLR can accept input $x_1$.**

Note that RLR requires the WFSA that it represents to match an entire input sequence (i.e., starting at the first symbol of the sequence). As an input sequence grows longer, it becomes exponentially more difficult for the WFSA to match the whole sequence. To circumvent this, we allow RLR to match a subsequence in the input sequence. This can be achieved by simply changing Equation 4 to

$$\mathbf{h}_{t+1} = \max\left((\mathbf{h}_t \, \mathbf{T}(x_{t+1})) \, \mathbf{T}(\epsilon)^*, \mathbf{h}_0\right) \approx \max((\mathbf{h}_t \, \mathbf{T}(x_{t+1})) \max(\mathbf{I}, \mathbf{T}(\epsilon)), \mathbf{h}_0). \tag{11}$$

By including $\mathbf{h}_0$ in the max function, RLR can begin the recurrence relation at time $t+1$, and start matching a subsequence beginning at time $t+1$ if it gives a higher score.

Rather than using only a single WFSA, our RLR model can include multiple WFSAs, with the number of WFSAs and their lengths (i.e., number of states) as hyperparameters. We term each WFSA in our RLR model as a *pattern* because a WFSA represents a sequential regularity in the input. We use the terms *WFSA* and *pattern* synonymously. (Figure 4 illustrates an example RLR model with 2 patterns, one of length 3 and another of length 4.) We stack the $p_{span}(\mathbf{x}) = \mathbf{h}_n \eta$ scores (Equations 5 and 11) of all patterns into a vector (where $n$ is the length of an input sequence), and pass that vector as input to a multilayer perceptron (MLP).

To learn the parameters of our RLR model, i.e., the weights and biases of its constituent WFSAs (the parameters in Equations 6 and 7) and the parameters of the MLP, we minimize cross-entropy and train the MLP and WFSAs end-to-end via backpropagation. To reduce the number of patterns used, we group all the weights and biases associated with each pattern, and penalizes our model using the $\mathcal{L}^2$ norm on the weights and biases.

In our healthcare application domain, a sequence of inputs $\mathbf{x} = x_1, \ldots, x_n$ corresponds to a sequence of clinical events (ClinicalEvents$_0$, ..., ClinicalEvents$_n$), where each ClinicalEvent$_i$ is a multi-dimensional vector. Each element in the vector contains the (integer or real) value of a clinical feature of a patient. An example of a 3-dimensional vector is ClinicalEvent$_i$=(Weight$_i$, AlbuminLevel$_i$, GlimepiridePrescribed$_i$).

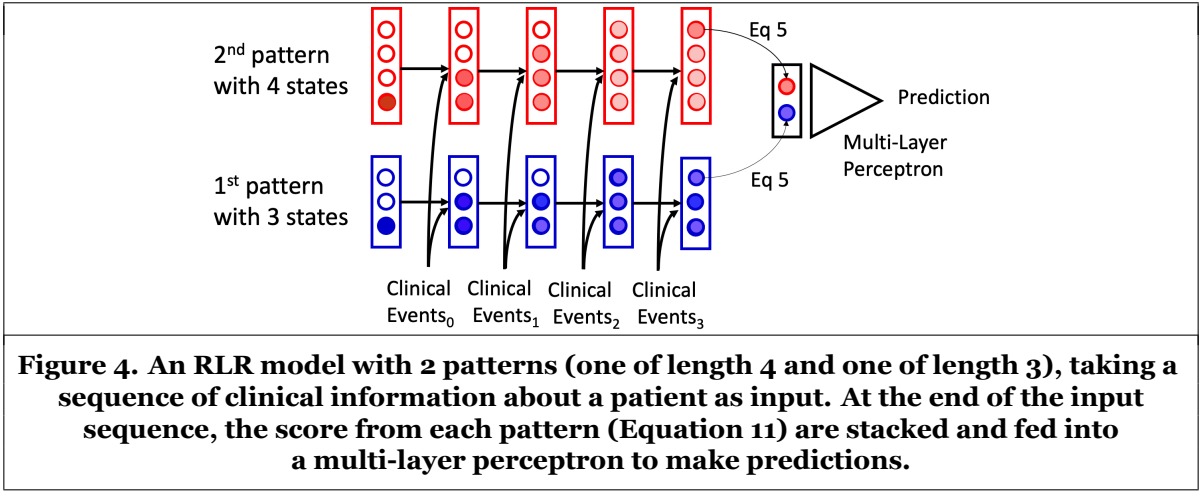

**Figure 4. An RLR model with 2 patterns (one of length 4 and one of length 3), taking a sequence of clinical information about a patient as input. At the end of the input sequence, the score from each pattern (Equation 11) are stacked and fed into a multi-layer perceptron to make predictions.**

## Experiments

### Datasets

We use two real-world clinical datasets (one public and one private) for four predictive tasks. The clinical events to be predicted range from short-term (within 24 hours) to long-term (within 1 year). Table 1 provides details about the data sizes, and the train/validation/test splits for each task.

#### MIMIC-III Dataset

We use the publicly available Medical Information Mart for Intensive Care (MIMIC-III) dataset (Johnson et al. 2016). The dataset contains longitudinal information about clinical events and medical outcomes of 33,798 unique patients with a total of 42,276 ICU stays. Harutyunyan et al. 2019 used MIMIC-III to prepare data for several benchmark tasks. We adopt that paper's data preparation methodology, and focus on the benchmark tasks of in-hospital mortality prediction and decompensation prediction. To represent the clinical information about a patient, we create a 76-dimensional vector for every hour of his ICU stay, with each element in the vector corresponding to a clinical feature about the patient.

**In-hospital Mortality Prediction.** For this task, we have to predict whether a patient dies during his ICU stay or survives to be discharged from ICU. The problem is one of binary classification. The input data consists of clinical events associated with each patient within the first 48 hours of his ICU stay. The exclusion of ICU stays that are less than 48 hours reduced the data size from 42,276 ICU stays to 21,139.

**Decompensation Prediction.** For this task, we have to predict whether a patient dies within the next 24 hours from every hour of his ICU stay. Because every hour of an ICU stay generates an example, a long stay produces many examples. Altogether there are 3,431,622 examples. This task is also one of binary classification.

| Task | Train (+ve; -ve) | Validation (+ve; -ve) | Test (+ve; -ve) |
|---|---|---|---|
| In-hospital Mortality (MIMIC-III) | 14,681 (1,987; 12,694) | 3,222 (436; 2,786) | 3,236 (374; 2,862) |
| Decompensation (MIMIC-III) | 2,377,768 (49,261; 2,328,507) | 530,646 (11,752; 518,894) | 523,208 (9,683; 513,525) |
| Eye Complications (DCD) | 126,681 (16,070; 110,611) | 27,154 (3,460; 23,694) | 27,159 (3,446; 23,713) |
| Ischemic Stroke (DCD) | 125,545 (7,059; 118,486) | 26,905 (1,515; 25,390) | 26,912 (1,527; 25,385) |

**Table 1.  Data sizes and splits**
**+ve: the number of positive cases, -ve: the number of negative cases**

**Diabetes Comorbidities Dataset (DCD)**

DCD is a proprietary dataset from a government hospital. This dataset contains about 190,000 de-identified patients who are afflicted with type 2 diabetes, spans a duration of 8 years (2011-2018), and covers a comprehensive collection of patient-specific longitudinal clinical information (e.g., demographic details, physiological measurements, diagnoses, medications, laboratory tests, surgical procedures, and visit dates). We preprocess the clinical data into a 57,153-dimensional vector for each 10-day interval in every patient's record. Each element in the vector corresponds to a clinical feature describing its associated patient. For this dataset, we focus on the tasks of predicting the onset of two diabetic comorbidities, viz., eye complications and ischemic stroke. For a comorbidity, each 10-day interval (including all information preceding it) constitutes an example, and it is labeled as positive if its associated patient is diagnosed with the comorbidity within a year from the end date of that interval (otherwise, it is labeled as negative).

**Eye Complications.** Diabetes is a leading cause of eye complications (e.g., diabetic retinopathy and macular edema) that could potentially result in permanent blindness. The vast majority of patients who develop such eye complications do not exhibit symptoms until the very late stages, by which time it is usually too late for effective treatment. Being able to accurately predict the onset of eye complications has the potential of identifying at-risk patients for early remedial treatment.

**Ischemic Stroke.** Diabetes is a well-established risk factor for ischemic stroke (Chen et al. 2016), which in turn, can result in mortality and other adverse outcomes such as pneumonia. Being able to predict the future onset of ischemic stroke has the potential of identifying at-risk patients for whom preemptive medical measures could be administered.

### RLR Model and Baselines

We set the hyperparameters of our RLR model (and those of the baselines) by tuning on validation data. For in-hospital mortality prediction (MIMIC-III), we use two 10-state, two 20-state, and two 50-state patterns (WFSAs). For decompensation prediction (MIMIC-III), we use thirty 3-state, thirty 4-state, and thirty 5-state patterns. For eye complications prediction (DCD), we use six 2-state, six 5-state, six 10-state, five 20-state, and seven 40-state patterns. For ischemic stroke prediction (DCD), we use six 2-state, six 5-state, seven 10-state, seven 20-state, and eight 40-state patterns. We used multi-layer perceptrons (MLPs) of different depths across datasets. For both tasks on MIMIC-III, we use a single-layer MLP. For both tasks on DCD, we use a 2-layer MLP.

We compare our RLR model against logistic regression, RETAIN (Choi et al. 2016a) and Adacare (Ma et al. 2020). Because logistic regression is a special case of RLR, it is a natural baseline to demonstrate that RLR's generalization improves its empirical performance beyond that of logistic regression. (For logistic regression, we have to 'flatten' the longitudinal EHR data by computing summary statistics (e.g., minimum, maximum, average, and standard deviation) for each feature.) RETAIN and Adacare (see section on *Related Work*) are chosen as baselines because they are state-of-the-art systems that model longitudinal healthcare data, and offer some degree of interpretability.

## Results

Table 2 reports the performances of our RLR model and the baselines on the four aforementioned tasks. We compare the systems using 3 metrics on test data: area under the precision-recall curve (AUPRC), area under the receiver operating characteristic curve (AUROC), and the log-likelihood (LL) of test examples. The advantage of using LL is that it directly measures the quality of the probability estimates produced (each system predicts the probability that a test example is true). The advantage of the AUPRC and AUROC is that they are insensitive to the large number of true negatives (i.e., test examples that are false and predicted to be false). The LL is the average over all test examples in a dataset. AUPRC and AUROC are computed by varying the threshold LL above which a test example is predicted to be true. For all metrics, the larger the number, the better the performance.

| MIMIC-III | In-house Mortality Prediction | | | | Decompensation Prediction | | | |
|---|---|---|---|---|---|---|---|---|
| Model | Logistic Regression | RETAIN | Adacare | RLR (Ours) | Logistic Regression | RETAIN | Adacare | RLR (Ours) |
| AUPRC | 0.4720 | 0.4514 | 0.4729 | **0.4911** | 0.2132 | 0.2597 | 0.3029 | **0.3247**[**] |
| AUROC | 0.8450 | 0.8467 | 0.8521 | **0.8540** | 0.8700 | 0.8764 | 0.9009 | **0.9043** |
| LL | -0.2680 | -0.2692 | -0.2668 | **-0.2648** | -0.0708 | -0.0692 | -0.0686 | **-0.0682**[*] |
| **DCD** | **Eye Complication Prediction** | | | | **Ischemic Stroke Prediction** | | | |
| Model | Logistic Regression | RETAIN | Adacare | RLR (Ours) | Logistic Regression | RETAIN | Adacare | RLR (Ours) |
| AUPRC | 0.2523 | 0.2616 | 0.3177 | **0.4108**[**] | 0.1484 | 0.2488 | 0.2793 | **0.3401**[**] |
| AUROC | 0.6877 | 0.6810 | 0.6869 | **0.7295**[*] | 0.7254 | 0.7242 | 0.6800 | **0.7935**[*] |
| LL | -0.6518 | -0.7236 | -0.3545 | **-0.3258**[*] | -0.6516 | -0.3372 | -0.1941 | **-0.1794**[*] |
| * and ** indicate statistical significance at the 95% and 99% level respectively. | | | | | | | | |

**Table 2. AUPRC, AUROC, LL (Log-likelihood) Results.**

From Table 2, we see the our RLR model is the best performer on all metrics for all four tasks.

Comparing the performance of RLR and logistic regression, we see that by generalizing beyond the latter, RLR achieves better performances on all metrics. Specifically, in terms of AUPRC, RLR outperforms logistic regression by 4% on mortality prediction, 52% on decompensation prediction, 63% on eye complications prediction, and 129% on ischemic stroke prediction.

Comparing the baselines RETAIN and Adacare, we see that Adacare is the stronger performer that consistently achieves better results. Comparing our RLR model to (the strongest baseline) Adacare, we see that RLR performs better than Adacare on all metrics for all tasks. Specifically, in terms of AUPRC, RLR outperforms Adacare by 4% on mortality prediction, 7% on decompensation prediction, 29% on eye complications prediction, and 22% on ischemic stroke prediction.

The better performance of RLR vis-a-vis RETAIN and Adacare can be explained by RLR being a simpler model. RLR can be viewed as simple chains of logistic regression models with such chains stacked one on top of another (Figure 4), whereas both RETAIN and Adacare are built from full-fledged complex recurrent neural networks. The simplicity of RLR allows it to use less data to train effectively, and hence RLR does better. This is especially useful in the clinical setting where positive examples are relatively scarce.

We conduct paired t-tests to evaluate the statistical significance of the test log-likelihood results. The null hypothesis is that a baseline is at least as good as RLR in terms of log-likelihood, and the alternative hypothesis is that RLR is better. For all tasks except mortality prediction, we are able to reject the null hypothesis, and support the alternative hypothesis that RLR is better than each baseline at the 95% confidence level. Note that the mortality prediction dataset is a relatively small one (in fact the smallest among the four datasets), and hence it is hard to establish statistical significance for it. Similarly, we show that most of the AUPRC and AUROC results for decompensation prediction, eye complications prediction, and ischemic stroke prediction are statistically significant at the 95% or 99% confidence level.[4]

**Performance on Small Datasets**

As mentioned in the *Introduction* section, RLR inherits the inductive bias from logistic regression, and hence would require less data to train than a deep learning system without such a bias. We verify RLR's performance in small data regimes by down-sampling the training data to about 5% for the in-hospital mortality prediction task, and to about 10% for the decompensation prediction task. The sizes of the test data are unchanged. Table 3 contains details of the down-sampled data sizes (we prefix the task names with 'Mini' to indicate that the tasks operate on small data sizes).

| Task | Train (+ve; -ve) | Validation (+ve; -ve) |
|---|---|---|
| Mini In-house Mortality (MIMIC-III) | 720 (94; 626) | 160 (25; 135) |
| Mini Decompensation (MIMIC-III) | 236,588 (4,995; 231,593) | 57,341 (1,207; 56,134) |

**Table 3. Down-Sampled Data Sizes**

Table 4 contains the performances of RLR and the baseline systems on test data after each is trained on the down-sampled training data. As before, the models are evaluated using the three metrics of AUPRC, AUROC, and LL. Compared to the deep learning systems Adacare and RETAIN, RLR continues to perform better on all metrics. Indeed RLR outperforms the deep learning systems by larger margins in the small data regime. In terms of AUPRC, on the decompensation prediction task, RLR outperforms Adacare by 17% (it is 7% formerly in Table 2 without down-sampling) and RETAIN by 63% (it is 25% formerly in Table 2). Likewise, on the in-house mortality prediction task, RLR outperforms Adacare by 9% (it is 4% formerly in Table 2) and RETAIN by 16% (it is 9% formerly in Table 2). The same trends can be observed for the other metrics. Compared to logistic regression in the small data regime, RLR still outperforms it on average, and loses to it only in terms of AUROC on the in-house mortality prediction task (but not statistically significantly). We observe that the margin of RLR's (better) performance over logistic regression shrinks in the small data regime. For example, on the decompensation prediction task, in terms of AUPRC, RLR outperforms logistic regression by 31% as opposed to 52% formerly in Table 2 without down-sampling. Other metrics behave similarly. The simplicity of logistic regression predisposes it to perform well in a small data regime where there is less data available to fit complex models with more parameters. Even so, RLR retains its competitiveness against logistic regression as the empirical results attest.

## *Interpretability*

To find out the relative contributions among the $k$ patterns of an RLR model to a prediction, we use a leave-one-out approach. Given the input sequence from a patient, RLR produces $k$ scores (Equations 5 and 11), one from each pattern. These scores are fed to a multi-layer perceptron (MLP; see Figure 4) to produce a (real-valued) prediction. After zeroing-out the score of a pattern (effectively holding out its input to the MLP), we note the change in the value produced by the MLP. The absolute value of this change reflects the contribution of the pattern to RLR's prediction. The $k$ patterns can thus be ranked in decreasing order of their associated (absolute) changes, with the top-ranked pattern being the one that contributes the most to the prediction.

---

[4]We ran experiments using 5-fold cross-validation on both MIMIC-III datasets, and obtained similar results, from which we can draw the same conclusions. However, due to space constraints, we left those numerical results out of the paper.

| MIMIC-III | Mini In-house Mortality Prediction | | | | Mini Decompensation Prediction | | | |
|---|---|---|---|---|---|---|---|---|
| Model | Logistic Regression | RETAIN | Adacare | RLR (Ours) | Logistic Regression | RETAIN | Adacare | RLR (Ours) |
| AUPRC | 0.4084 | 0.3610 | 0.3806 | **0.4170** | 0.1713 | 0.1380 | 0.1921 | **0.2249**[**] |
| AUROC | **0.8305** | 0.7856 | 0.7944 | 0.8067 | 0.8418 | 0.8154 | 0.8628 | **0.8681** |
| LL | -0.2993 | -0.3143 | -0.3316 | **-0.2908** | -0.0800 | -0.0891 | -0.0869 | **-0.0779**[**] |
| * and ** indicate statistical significance at the 95% and 99% level respectively. | | | | | | | | |

**Table 4. AUPRC, AUROC, LL (Log-likelihood) Results**
**(with Down-Sampled Training Data)**

We investigate the interpretability of the patterns (WFSAs) learned by our RLR model at the individual patient level and at the (sub)population level.

At the individual level, we pick a patient $P$ who dies in ICU from the validation set[5] of the decompensation task (i.e., the patient is a positive example for the task).

The top pattern for patient $P$ is shown in Figure 5(a). This pattern has 4 states. This pattern is activated at time 376 (a time step correspond to an hour). $P$ moves from state 0 to 1 because of a Glasgow Coma Scale (GCS) Total score of 4. (Recall from the *RLR* section that each transition corresponds to a logistic regression. Hence we can find the top input features for that transition by identifying those that sum up to more than 80% of the log-odds of the logistic regression. This is how we identified GCS Total as an important feature for the first transition.) Next, $P$ moves from state 1 to 2 at time 377 because of (again) a GCS Total score of 4. $P$ then loops and stays in state 2 until time 380, at which point it moves to state 3 because of a lower GCS Total score of 3. $P$ loops for 18 time steps and stays at state 3 till time 398. At that point, it transitions to state 4 because of the same GCS Total score of 3, and low GCS Motor Response and Verbal Response scores of 1. This pattern captures the gradual degradation of the patient through time as evidenced by the general trend of decreasing GCS scores. This pattern aligns with medical knowledge. The Glasgow Coma Scale (GCS) is a clinical scale used to measure a person's level of consciousness after a brain injury. The GCS assesses a person based on his ability to move his eyes and body, and to speak. Lower GCS scores are correlated with a higher risk of death. Figures 5(b), (c), (d), and (e) respectively plot the scores for $P$'s GCS Verbal Response, Total, Eye Opening, and Motor Response across time. The plots show that the pattern is capturing the last snippet on patient $P$'s timeline, and show that the pattern is combining the four GCS scores to capture a general trend of decline. First, note that by focusing on the end of the timeline, the pattern is correctly focusing on the period critical to the patient's death. Second, note that the transitions from state to state in the pattern occurs at different intervals; the pattern allows this through the use of self loops. This is an illustration of how RLR is able to model multi-time-scale events.

The pattern in Figure 5(a) illustrates how RLR provides interpretability by threading together salient features into a coherent whole (previously mentioned in the *Related Work* section). This is in sharp contrast to the attention mechanism of a deep learning system that typically highlights important features across the entire timespan without clarifying how they relate to one another. For example, in Figure 5(a), the important features are GCS Total score (at time 376, 377, 380, 398), Eye Opening score (at time 377), GCS Verbal Response score (at time 398), and GCS Motor Response score (at time 398). The attention mecha-nism merely highlights all these features as important without discerning whether all of them should form one regular sequence, or whether they should be broken up into subsets, each of which is pieced together to form a different sequence, such as:

1. GCS Total score$_{t=376}$ → GCS Total score$_{t=377}$

2. GCS Eye Opening$_{t=377}$ → GCS Verbal Response$_{t=398}$, GCS Motor Response$_{t=398}$

---

[5]We do not use the test set to prevent us from inadvertently exposing our RLR model to the test examples, and thereby compromising the veracity of the empirical results in Table 2.

RLR does not have such a problem. Indeed, RLR provides a perspicuous view of how the features are linked together temporally in distinct sequences, and does not leave end-users wondering which combination of features go together in a pattern.

Next we repeat the analysis for a patient $Q$ from the validation set of the eye complications prediction task. Figure 6(a) illustrates the most important pattern for $Q$. $Q$ moves from state 0 to 1 using an $\epsilon$-transition, i.e., the input vector at that time point is ignored. $Q$ then moves from state 1 to 2 because of his age, high creatinine levels, and history of diabetes. It then loops and stays at state 2 for 130 time steps, and then moves via an $\epsilon$-transition to state 3. At time 154, $Q$ moves from state 3 to state 4. because of a high age value, and a high creatinine level. This WFSA shows that a pattern of persistently high creatinine at an advanced age, coupled with a history of diabetes, is predictive of eye complications. This is consistent with medical research (Rajalakshmi et al. 2020; Zhang et al. 2019) that finds diabetic retinopathy to be positively correlated with age, duration of diabetes, and creatinine levels.

To find the relative contributions of RLR's patterns at the (sub)population level, we simply repeat the leave-one-out approach for every patient, and find the average (absolute) change in prediction value of each pattern across all patients. The pattern with the largest change in value is the one that contributes the most to RLR's predictions on average. Using this process, we identify the most important pattern on decompensation's validation set (Figure 7). The pattern sensibly captures the intuition that a low systolic blood pressure, followed by a low heart rate and a low respiratory rate, is predictive of patient's death.

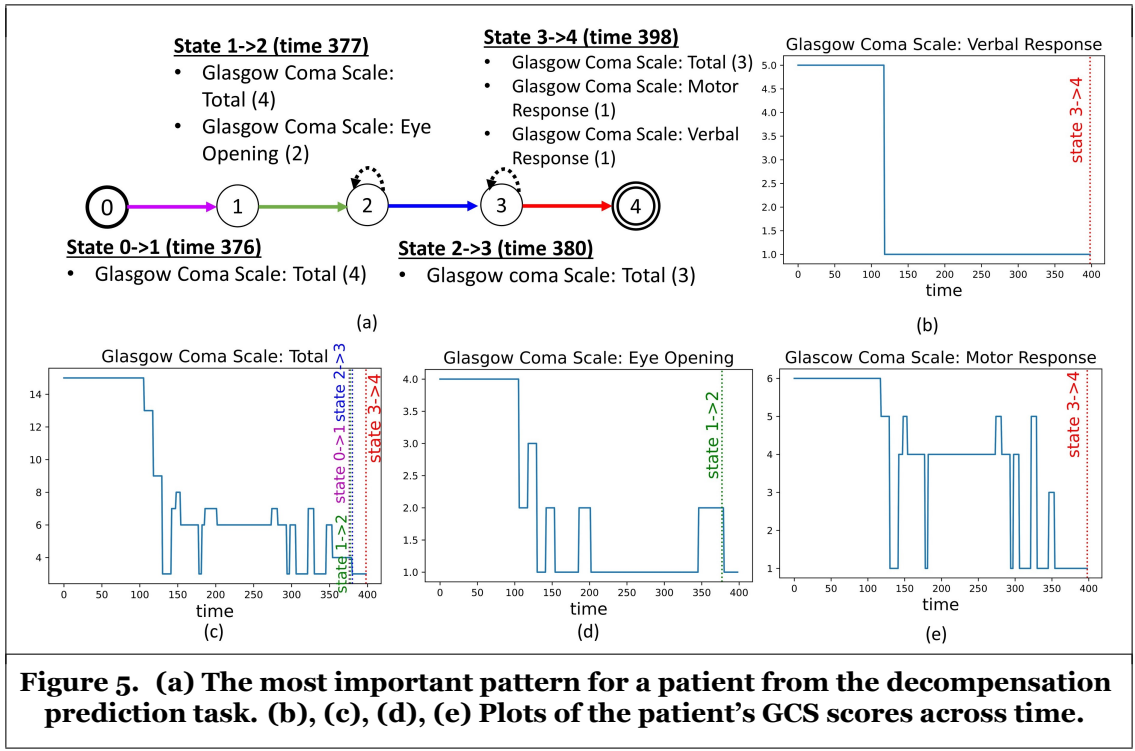

**Figure 5. (a) The most important pattern for a patient from the decompensation prediction task. (b), (c), (d), (e) Plots of the patient's GCS scores across time.**

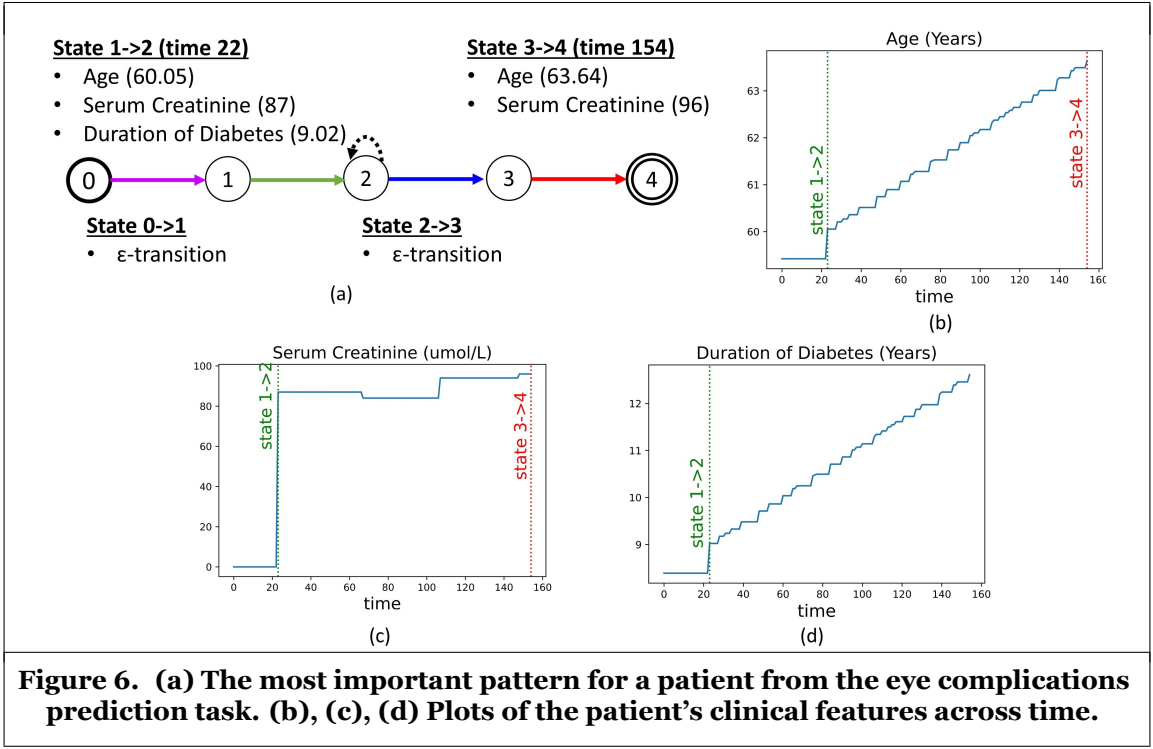

**Figure 6.** (a) The most important pattern for a patient from the eye complications prediction task. (b), (c), (d) Plots of the patient's clinical features across time.

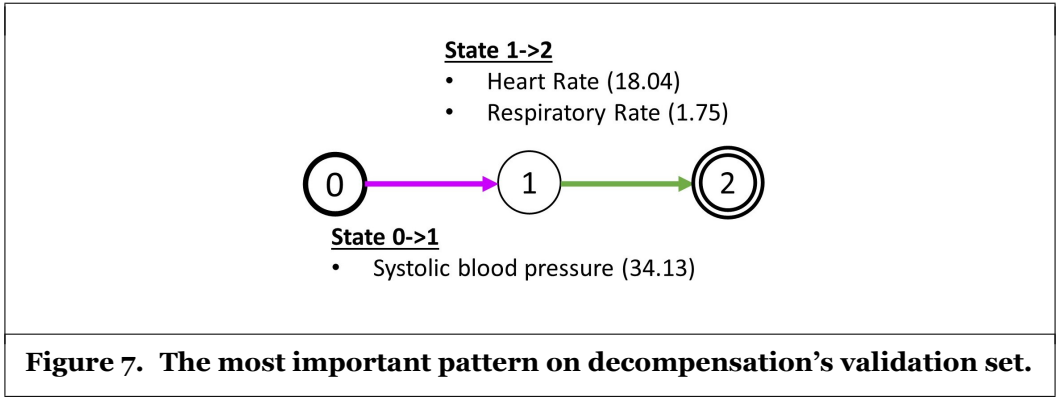

**Figure 7.** The most important pattern on decompensation's validation set.

## Conclusion and Future Work

In this paper, we present RLR, a novel model that predicts clinical events from EHR data, and learns interpretable patterns as weighted finite state automata (WFSAs). We show that RLR is a generalization of logistic regression to longitudinal time-series data. RLR chains multiple logistic regression models together, and stacks these chains one on top of another. The self-loops and ε-transitions in the WFSAs underlying RLR allow it to flexibly learn patterns across different time-scales. Experimental comparisons of RLR against logistic regression and two state-of-the-art models on four real-world tasks demonstrate the promise of our approach. Empirical analysis of RLR's learned patterns shows the meaningfulness of the knowledge that is captured by the patterns. In terms of interpretability, RLR surpasses existing methods by chaining together features into coherent, distinct sequential patterns. This is in contrast to extant methods such as attention mechanisms of deep learning systems that merely highlight important features across an entire timespan without explicating the possibly multifarious sequential structures interweaving among them. As future work, we want to use evolutionary algorithms to search for the number of WFSAs and their lengths, rather than specifying these as hyperparameters. We also want to move beyond WFSAs, and incorporate more powerful automata higher in the Chomsky hierarchy (Chomsky 1956) into RLR.

## Acknowledgements

This research is partly supported by Singapore Ministry of Health's National Innovation Challenge Grant (MOH/NIC/CDM1/2018) and Singapore Ministry of Education's AcRF Tier 1 Grant (R-253-000-146-133) to Stanley Kok. Any opinions, findings, conclusions, or recommendations expressed in this material are those of the authors, and do not reflect the views of the funding agencies.